\documentclass{article}

\usepackage[final,nonatbib]{assets/techreport}

\usepackage[utf8]{inputenc}
\usepackage[T1]{fontenc}
\usepackage{hyperref}
\usepackage{url}
\usepackage{booktabs}
\usepackage{amsfonts}
\usepackage{nicefrac}
\usepackage{microtype}
\usepackage{xcolor}
\usepackage{pgfplots}
\pgfplotsset{compat=1.17}
\usepackage{subcaption}
\usepackage{csvsimple}
\usepackage{siunitx}
\usepackage[flushleft]{threeparttable}
\usepackage{enumitem}
\usepackage{etoolbox}
\usepackage{pgfmath}
\usetikzlibrary{matrix}
\usetikzlibrary{calc}
\usepgfplotslibrary{groupplots}

\title{VitalLens 2.0: High-Fidelity rPPG for Heart Rate Variability Estimation from Face Video}

\author{%
  Philipp V.~Rouast \\
  Rouast Labs\\
  \texttt{philipp@rouast.com} \\
}

\csvreader[before reading=\def\nparticipantstrain{0}\def\nchunkstrain{0}\def\totaltimetrain{0}]
    {data/training_summary.csv}
    {Participants=\participants,Chunks=\chunks,Time=\time}
    {%
        \pgfmathsetmacro{\nparticipantstrain}{\nparticipantstrain+\participants}%
        \pgfmathsetmacro{\nchunkstrain}{\nchunkstrain+\chunks}%
        \pgfmathsetmacro{\totaltimetrain}{\totaltimetrain+\time}%
    }

\csvreader[before reading=\def\nparticipantstest{0}\def\nchunkstest{0}]
    {data/testset_summary.csv}
    {participants=\participants, chunks=\chunks}
    {%
        \pgfmathsetmacro{\nparticipantstest}{\nparticipantstest+\participants}%
        \pgfmathsetmacro{\nchunkstest}{\nchunkstest+\chunks}%
    }

\csvreader[head to column names]
    {data/best_values.csv}
    {metric=\metric, best_value=\mybestval}
    {
      \expandafter\pgfmathsetmacro\csname best\metric\endcsname{\mybestval}
    }%

\csvreader[head to column names]
    {data/vl2_values.csv}
    {metric=\metric, val=\myval}
    {
      \expandafter\pgfmathsetmacro\csname vl\metric\endcsname{\myval}
    }%

\newcommand{\roundnumdigits}[2]{\num[round-mode=places,round-precision=#1]{#2}}
\newcommand{\printval}[2]{%
  \ifdim #1 pt = 0.0 pt
    --%
  \else
    \ifdim #1 pt = \csname #2\endcsname pt
      \textbf{\num{#1}}%
    \else
      \num{#1}%
    \fi
  \fi
}

\newcommand{\printnum}[1]{%
  \ifdim #1 pt = 0.0 pt
    --%
  \else
    \num{#1}%
  \fi
}

\begin{document}
\begin{filecontents*}{data/training_summary.csv}
Source,Participants,Chunks,Time
PROSIT (In-house),128.0,5700.0,16.8
Vital Videos (EU),589.0,2238.0,9.3
Vital Videos (Africa),383.0,2869.0,7.7
Vital Videos (South Asia),313.0,3887.0,10.8
\end{filecontents*}

\begin{filecontents*}{data/testset_summary.csv}
source_dataset,participants,chunks
PROSIT (In-house),22,251
Vital Videos (EU),146,282
Vital Videos (Africa),93,260
Vital Videos (South Asia),78,168
UBFC-Phys,34,67
UBFC-rPPG,49,53
\end{filecontents*}

\begin{filecontents*}{data/best_values.csv}
metric,best_value
hrmae,1.56716
hrppgcor,0.85688
hrppgsnr,13.51550
rrmae,1.08427
rrrespcor,0.81616
rrrespsnr,10.24858
hrvsdnnmae,10.18038
hrvrmssdmae,16.44912
hrvlfhfmae,0.83443
\end{filecontents*}

\begin{filecontents*}{data/vl2_values.csv}
metric,val
hrmae,1.56716
hrppgcor,0.85688
hrppgsnr,13.51550
rrmae,1.08427
rrrespcor,0.81616
rrrespsnr,10.24858
hrvsdnnmae,10.18038
hrvrmssdmae,16.44912
hrvlfhfmae,0.83443
\end{filecontents*}

\maketitle

\begin{abstract}
This report introduces VitalLens 2.0, a new deep learning model for estimating physiological signals from face video. This new model demonstrates a significant leap in accuracy for remote photoplethysmography (rPPG), enabling the robust estimation of not only heart rate (HR) and respiratory rate (RR) but also Heart Rate Variability (HRV) metrics. This advance is achieved through a combination of a new model architecture and a substantial increase in the size and diversity of our training data, now totaling \roundnumdigits{0}{\nparticipantstrain} unique individuals. We evaluate VitalLens 2.0 on a new, combined test set of \roundnumdigits{0}{\nparticipantstest} unique individuals from four public and private datasets. When averaging results by individual, VitalLens 2.0 achieves a Mean Absolute Error (MAE) of \roundnumdigits{2}{\vlhrmae} bpm for HR, \roundnumdigits{2}{\vlrrmae} bpm for RR, \roundnumdigits{2}{\vlhrvsdnnmae} ms for HRV-SDNN, and \roundnumdigits{2}{\vlhrvrmssdmae} ms for HRV-RMSSD. These results represent a new state-of-the-art, significantly outperforming previous methods. This model is now available to developers via the VitalLens API at \textbf{\url{https://rouast.com/api}}.
\end{abstract}

\section{Introduction}
\label{sec:introduction}

Remote Photoplethysmography (rPPG) harnesses signals from standard video to estimate physiological information, offering immense potential for non-invasive health monitoring \cite{verkruysse2008remote}. Our initial release, VitalLens 1.0, provided real-time estimation of heart rate (HR) and respiratory rate (RR) \cite{rouast2023vitallens}. 

This report introduces VitalLens 2.0, the next generation of our rPPG model. The primary goal is to move beyond simple rate estimation to achieve high-fidelity physiological waveform reconstruction. This moves the challenge from simple rate estimation (i.e., finding a dominant frequency) to high-fidelity waveform reconstruction, which requires the precise, sub-second temporal accuracy of Inter-Beat Intervals (IBIs) to be robust.

This leap in performance was achieved through these two key developments:
\begin{enumerate}
    \item \textbf{An expanded and meticulously curated training dataset.} We combined our in-house data with the Vital Videos  public dataset, then manually curated all samples to ensure high-quality video and labels. The final set totals \roundnumdigits{0}{\nparticipantstrain} unique individuals.
    \item \textbf{A new model architecture and training methodology}, designed specifically to capture the subtle inter-beat variations necessary for accurate HRV analysis.
\end{enumerate}

To validate these improvements, we establish a new, large-scale combined test set. This set is comprised of \roundnumdigits{0}{\nchunkstest} video samples from \roundnumdigits{0}{\nparticipantstest} unique individuals, combining our in-house test set (PROSIT) with several publicly available datasets (Vital Videos \cite{toye2023vital, toye2025rich}, UBFC-Phys \cite{sabour2021ubfc}, UBFC-rPPG \cite{bobbia2019unsupervised}). All video chunks are processed to be 20-60 seconds in duration, making them suitable for time-domain HRV analysis. For our model development purposes we created our own strict, participant-disjoint training, validation, and test sets to ensure a robust, unbiased evaluation of generalization.

Our key contributions are:
\begin{itemize}
    \item \textbf{VitalLens 2.0.} We introduce a new rPPG model capable of accurately estimating HRV metrics from face video, in addition to HR and RR.
    \item \textbf{Comprehensive evaluation.} We benchmark VitalLens 2.0 on a large, diverse test set of \roundnumdigits{0}{\nparticipantstest} individuals, demonstrating its superior accuracy.
    \item \textbf{State-of-the-art performance.} We show that VitalLens 2.0 significantly outperforms handcrafted algorithms (e.g., POS \cite{wang2017algorithmic}), other learning-based methods (e.g., MTTS-CAN \cite{liu2020multi}), and our previous VitalLens 1.0 architecture.
\end{itemize}

\vspace{1em}
This new model is the engine behind the VitalLens API, enabling developers to integrate robust, real-time HRV analysis into their applications. For more information, visit \textbf{\url{https://rouast.com/api}}.

\section{Architecture}
\label{sec:architecture}

The VitalLens 2.0 model is an end-to-end deep convolutional neural network, building upon an EfficientNet-based backbone \cite{tan2021efficient}. The architecture incorporates novel temporal-attentive mechanisms optimized for extracting high-fidelity physiological waveforms. This design is specifically focused on minimizing signal noise and preserving the precise temporal location of systolic peaks, which is the critical prerequisite for reliable IBI extraction and HRV analysis.

The model takes a sequence of video frames of a person's face as input and outputs two continuous time-series signals: the pulse (PPG) waveform and the respiration (RESP) waveform. From these waveforms, downstream metrics such as Heart Rate (HR), Respiratory Rate (RR), and HRV metrics (e.g., SDNN, RMSSD, LF/HF) are derived using industry-standard signal processing techniques.

\begin{figure}[h!]
    \centering

    \begin{tikzpicture}
        \matrix [
            matrix of nodes, 
            anchor=south, 
            column sep=0.5em,
            row sep=0pt,
            nodes={font=\small, anchor=west, inner sep=1pt},
            cells={anchor=west}
        ]
        {
            \draw [black, thick] (0,0.1em) -- (0.3cm, 0.1em); & \node {Ground Truth}; &
            \draw [red, thick] (0,0.1em) -- (0.3cm, 0.1em);   & \node {VitalLens 2.0 (PPG)}; &
            \draw [blue, thick] (0,0.1em) -- (0.3cm, 0.1em);  & \node {VitalLens 2.0 (RESP)}; &
            \draw [cyan, thick] (0,0.1em) -- (0.3cm, 0.1em);& \node {VitalLens 1.0 (PPG)}; &
            \draw [green, thick] (0,0.1em) -- (0.3cm, 0.1em); & \node {POS (PPG)}; \\
        };
    \end{tikzpicture}

    \begin{subfigure}{\textwidth}
        \centering
        \begin{tikzpicture}
            \begin{axis}[
                height=3.6cm, width=\textwidth,
                ylabel={PPG},
                xticklabels={}
            ]
            \addplot[black, thick, opacity=0.8] table[col sep=comma, x=time, y=gt_ppg] {data/waveform_comparison.csv};
            \addplot[red, thick] table[col sep=comma, x=time, y=vl2_ppg] {data/waveform_comparison.csv};
            \end{axis}
        \end{tikzpicture}
        \label{subfig:wf-vl2-ppg}
    \end{subfigure}

    \begin{subfigure}{\textwidth}
        \centering
        \begin{tikzpicture}
            \begin{axis}[
                height=3.6cm, width=\textwidth,
                ylabel={RESP},
                xticklabels={}
            ]
            \addplot[black, thick, opacity=0.8] table[col sep=comma, x=time, y=gt_resp] {data/waveform_comparison.csv};
            \addplot[blue, thick] table[col sep=comma, x=time, y=vl2_resp] {data/waveform_comparison.csv};
            \end{axis}
        \end{tikzpicture}
        \label{subfig:wf-vl2-resp}
    \end{subfigure}

    \begin{subfigure}{\textwidth}
        \centering
        \begin{tikzpicture}
            \begin{axis}[
                height=3.6cm, width=\textwidth,
                ylabel={PPG},
                xlabel={Time (seconds)}
            ]
            \addplot[black, thick, opacity=0.8] table[col sep=comma, x=time, y=gt_ppg] {data/waveform_comparison.csv};
            \addplot[cyan, thick] table[col sep=comma, x=time, y=vl1_ppg] {data/waveform_comparison.csv};
            \addplot[green, thick] table[col sep=comma, x=time, y=pos_ppg] {data/waveform_comparison.csv};
            \end{axis}
        \end{tikzpicture}
        \label{subfig:wf-others}
    \end{subfigure}
    
    \caption{Visual comparison of estimated waveforms from a sample handheld video segment.\protect\footnotemark{} Top: VitalLens 2.0 PPG vs. Ground Truth. Middle: VitalLens 2.0 Respiration vs. Ground Truth. Bottom: VitalLens 1.0 and POS PPG vs. Ground Truth. VitalLens 2.0 achieves higher fidelity, accurately reconstructing the precise timing of the systolic peaks.}
    \label{fig:waveform-fidelity}
\end{figure}
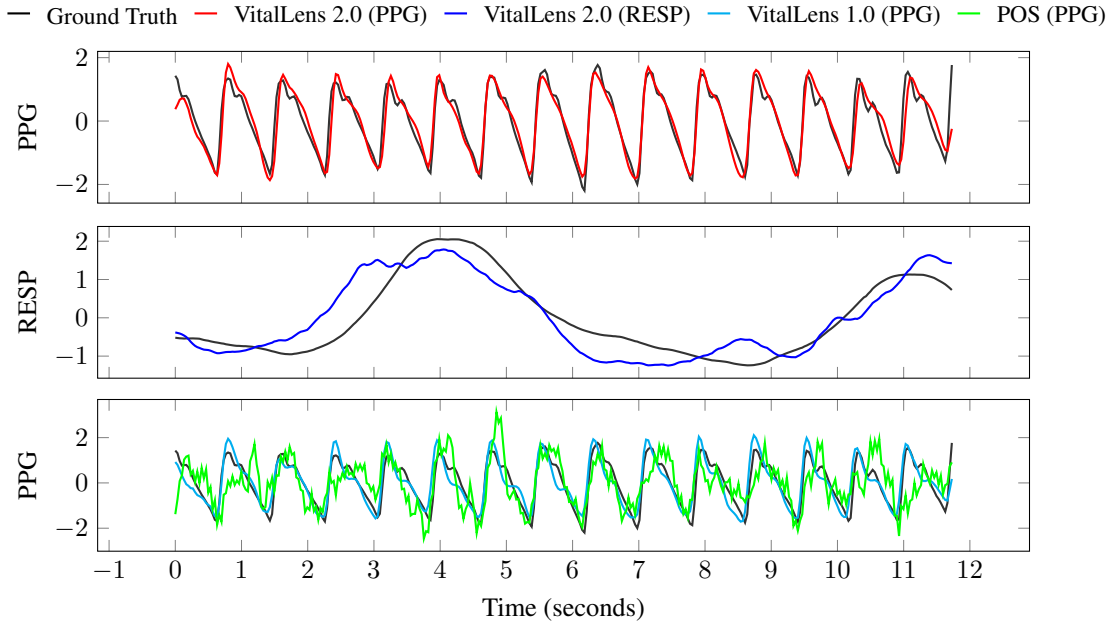

Figure \ref{fig:waveform-fidelity} provides a visual comparison of the high-fidelity waveforms generated by VitalLens 2.0 against the ground truth and other models for a sample handheld video segment.

\footnotetext{The source video and ground truth vitals are publicly available as \texttt{sample\_video\_1} at: \url{https://github.com/Rouast-Labs/vitallens-python/tree/main/examples}}

\section{Datasets}
\label{sec:datasets}

The development of VitalLens 2.0 utilized a large-scale, multi-source dataset, combining our in-house PROSIT dataset with several publicly available datasets, including the Vital Videos (VV) collection \cite{toye2023vital, toye2025rich}, UBFC-Phys \cite{sabour2021ubfc}, and UBFC-rPPG \cite{bobbia2019unsupervised}.

\paragraph{Training Dataset.}
The training dataset combines the training splits of PROSIT and the Vital Videos datasets we have access to.
This results in a total training set of \roundnumdigits{0}{\nparticipantstrain} unique individuals, a significant increase in size and diversity over the data used for VitalLens 1.0 \cite{rouast2023vitallens}. Beyond demographics, this diversity includes a wide variety of real-world filming locations, diverse lighting conditions, cameras, and varied backgrounds, and unscripted participant behavior, including significant camera motion from handheld devices. The composition of this training set is detailed in Table \ref{tab:training-summary}.

\begin{table}[h!]
    \caption{VitalLens 2.0 Training Dataset Size}
    \label{tab:training-summary}
    \centering
    \begin{tabular}{lrrr}
    	\toprule
    	Source & \# Participants & \# Chunks & Time (hours) \\
    	\midrule
    	\csvreader[late after line=\\]
        {data/training_summary.csv}
        {Source=\source, Participants=\participants, Chunks=\chunks, Time=\time}
        {
         \source & \roundnumdigits{0}{\participants} & \roundnumdigits{0}{\chunks} & \roundnumdigits{1}{\time}%
        }
        \midrule
    	\textbf{Total} & \textbf{\roundnumdigits{0}{\nparticipantstrain}} & \textbf{\roundnumdigits{0}{\nchunkstrain}} & \textbf{\roundnumdigits{1}{\totaltimetrain}} \\
    	\bottomrule
    \end{tabular}
\end{table}

\definecolorseries{myseries1}{rgb}{step}[rgb]{1,0,0}{-1,0,1}
\resetcolorseries{myseries1}%
\definecolorseries{myseries2}{rgb}{step}[rgb]{1.0, 0.75, 0.65}{-0.134,-0.122,-0.13}
\resetcolorseries{myseries2}%
\newcommand{\slice}[5]{
    \pgfmathsetmacro{\midangle}{0.5*#1+0.5*#2}
    \begin{scope}
        \clip (0,0) -- (#1:1) arc (#1:#2:1) -- cycle;
        \colorlet{SliceColor}{#3!!+}
        \fill[inner color=SliceColor!30,outer color=SliceColor!60] (0,0) circle (1cm);
    \end{scope}
    \draw[thick] (0,0) -- (#1:1) arc (#1:#2:1) -- cycle;
    \node[label={[font=\small]\midangle:#5}] at (\midangle:1) {};
    \pgfmathsetmacro{\temp}{min((#2-#1-10)/110*(-0.3),0)}
    \pgfmathsetmacro{\innerpos}{max(\temp,-0.5) + 0.8}
    \node[font=\small] at (\midangle:\innerpos) {#4};
}

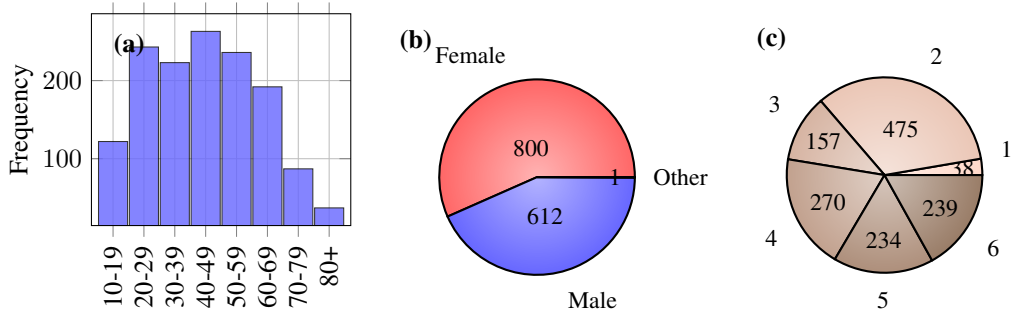
\begin{figure}[h!]
    \centering
    \begin{subfigure}{0.36\textwidth}
        \centering
        \begin{tikzpicture}
            \begin{axis}[
                    table/col sep=comma,
                    ybar,
                    ylabel=Frequency,
                    width=\textwidth,
                    height=4.4cm,
                    grid=major,
                    bar width=11pt,
                    label style={inner sep=0pt},
                    xticklabel style={/pgf/number format/1000 sep=, rotate=90, anchor=east},
                    xticklabels from table={data/age_histogram.csv}{age_bin},
                    xtick=data
                ]
            \addplot[fill=blue!60,opacity=0.8] table[x expr=\coordindex, y=count] {data/age_histogram.csv};
            
            \node[anchor=north west, font=\bfseries] at (rel axis cs:0.05,0.95) {(a)};
            
            \end{axis}
        \end{tikzpicture}
    \end{subfigure}
    \hfill
    \begin{subfigure}{0.3\textwidth}
        \centering
        \begin{tikzpicture}[scale=1.3]%
            \def\mya{0}\def\myb{0}
            \begin{filecontents*}{data/gender.csv}
Label,Value
Female,800
Male,612
Other,1
\end{filecontents*}
\csvreader[before reading=\def\mysum{0}]{data/gender.csv}{Value=\Value}{%
                \pgfmathsetmacro{\mysum}{\mysum+\Value}%
            }
            \csvreader[head to column names]{data/gender.csv}{}{%
                \let\mya\myb
                \pgfmathsetmacro{\myb}{\myb+\Value}
                \slice{\mya/\mysum*360}{\myb/\mysum*360}{myseries1}{\Value}{\Label}
            }
            
            \node[anchor=north west, font=\bfseries] at (-1.5, 1.6) {(b)};

        \end{tikzpicture}%
    \end{subfigure}
    \hfill
    \begin{subfigure}{0.3\textwidth}
        \centering
        \begin{tikzpicture}[scale=1.3]%
            \begin{filecontents*}{data/skin_type.csv}
Label,Value
1,38
2,475
3,157
4,270
5,234
6,239
\end{filecontents*}
\csvreader[before reading=\def\mysum{0}]{data/skin_type.csv}{Value=\Value}{%
                \pgfmathsetmacro{\mysum}{\mysum+\Value}%
            }
            \def\mya{0}\def\myb{0}
            \csvreader[head to column names]{data/skin_type.csv}{}{%
                \let\mya\myb
                \pgfmathsetmacro{\myb}{\myb+\Value}
                \slice{\mya/\mysum*360}{\myb/\mysum*360}{myseries2}{\Value}{\Label}
            }
            
            \node[anchor=north west, font=\bfseries] at (-1.4, 1.6) {(c)};

        \end{tikzpicture}%
    \end{subfigure}
    
    \caption{Participant demographics in the training dataset. (a) Age, (b) Gender, (c) Skin type.}
    \label{fig:training-demographics}
\end{figure}

\begin{figure}[h!]
    \centering
    \begin{tikzpicture}
        \begin{groupplot}[
            group style={
                group size=3 by 1,
                horizontal sep=1.5em,
                vertical sep=1em
            },
            height=4.4cm,
            width=0.35\textwidth,
            table/col sep=comma,
            ybar,
            grid=major,
            xticklabel style={
                /pgf/number format/1000 sep=,
                rotate=90,
                anchor=east,
                font=\small
            },
            xtick=data,
            ymin=0
        ]

        \nextgroupplot[
            ylabel=Frequency,
            xlabel={Heart rate (bpm)},
            xticklabels from table={data/hr_histogram.csv}{bin}
        ]
        \addplot[fill=red!60, opacity=0.8] table[x expr=\coordindex, y=count] {data/hr_histogram.csv};
        \node[anchor=north west, font=\bfseries] at (rel axis cs:0.05,0.95) {(a)};

        \nextgroupplot[
            xlabel={Resp. rate (bpm)},
            xticklabels from table={data/rr_histogram.csv}{bin},
            yticklabels={}
        ]
        \addplot[fill=blue!60, opacity=0.8] table[x expr=\coordindex, y=count] {data/rr_histogram.csv};
        \node[anchor=north west, font=\bfseries] at (rel axis cs:0.05,0.95) {(b)};

        \nextgroupplot[
            xlabel={SDNN (ms)},
            xticklabels from table={data/sdnn_histogram.csv}{bin},
            yticklabels={}
        ]
        \addplot[fill=purple!60, opacity=0.8] table[x expr=\coordindex, y=count] {data/sdnn_histogram.csv};
        \node[anchor=north west, font=\bfseries] at (rel axis cs:0.05,0.95) {(c)};

        \end{groupplot}
    \end{tikzpicture}
    \caption{Distributions of per-individual average vitals in the combined training dataset. (a) Heart Rate, (b) Respiratory Rate, (c) HRV-SDNN.}
    \label{fig:training-vitals}
\end{figure}
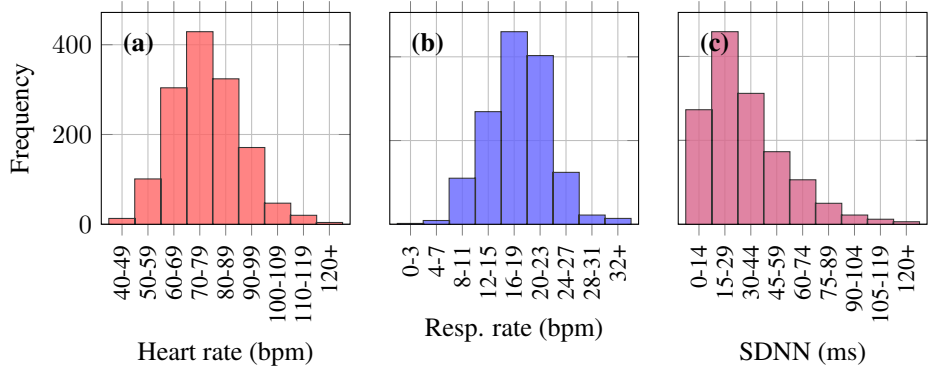

Figures \ref{fig:training-demographics} and \ref{fig:training-vitals} detail the demographic and physiological composition of this new training set. Data is reported per-individual. It is well-balanced in gender and includes a comprehensive representation across all six Fitzpatrick skin types. The vitals distributions cover a wide range of physiological states, including a broad spectrum of heart rate variability.

\paragraph{Test Dataset.}
To validate VitalLens 2.0, we constructed a new, combined test set. All individuals in this test set are disjoint from the training and validation sets. Chunks were re-processed to have durations between 20 and 60 seconds, enabling the calculation of time- and frequency-domain HRV metrics. The composition of this test set is detailed in Table \ref{tab:testset-summary}.
While all test samples have a ground truth signal for PPG, some (VV in part and both UBFC-rPPG and UBFC-Phys) do not have a ground truth signal for RESP.
In addition, samples with known label-quality or synchronization issues (e.g., in parts of UBFC-Phys) were excluded to ensure a fair and reliable benchmark.

\begin{table}[h!]
  \caption{VitalLens 2.0 Combined Test Set Composition}
  \label{tab:testset-summary}
  \centering
  \begin{tabular}{lrr}
    \toprule
    Source Dataset & \# Participants & \# Chunks \\
    \midrule
    \csvreader[late after line=\\]
        {data/testset_summary.csv}
        {source_dataset=\sourceDataset, participants=\participants, chunks=\chunks}
        {
         \sourceDataset & \roundnumdigits{0}{\participants} & \roundnumdigits{0}{\chunks}%
        }
    \midrule
    \textbf{Total} & \textbf{\roundnumdigits{0}{\nparticipantstest}} & \textbf{\roundnumdigits{0}{\nchunkstest}} \\
    \bottomrule
  \end{tabular}
\end{table}

\section{Methodology}
\label{sec:methodology}

\paragraph{Model Training and Validation.}
We follow a strict participant-disjoint methodology for training, validation, and testing. The \roundnumdigits{0}{\nparticipantstrain} individuals in the training set were used to optimize model parameters. A separate validation set, also participant-disjoint from the training set, was used to monitor for overfitting and to perform hyperparameter tuning and model selection. The final VitalLens 2.0 model chosen for evaluation is the one that demonstrated the best performance on this validation set.

\paragraph{Models Compared.}
We benchmark VitalLens 2.0 against a comprehensive set of methods. These include traditional handcrafted algorithms (G \cite{verkruysse2008remote}, CHROM \cite{de2013robust}, and POS \cite{wang2017algorithmic}) and several prominent learning-based methods (DeepPhys \cite{chen2018deep}, MTTS-CAN \cite{liu2020multi}, and EfficientPhys \cite{liu2023efficientphys}). To ensure a fair and direct comparison, all learning-based baselines were re-trained from scratch on the exact same \roundnumdigits{0}{\nparticipantstrain}-participant training dataset used for VitalLens 2.0.

We also include two internal baselines:
\begin{itemize}
    \item \textbf{VitalLens 1.0*:} The original model as presented in \cite{rouast2023vitallens}, trained on the smaller, original dataset.
    \item \textbf{VitalLens 1.1:} The original VitalLens 1.0 architecture re-trained on our new, larger \roundnumdigits{0}{\nparticipantstrain}-participant dataset.
\end{itemize}
This comparison allows us to isolate performance gains from architectural improvements (VitalLens 2.0 vs. 1.1) versus data improvements (VitalLens 1.1 vs. other baselines and 1.0*).

\paragraph{Evaluation Metrics.}
We evaluate performance at both the waveform and vital sign levels. For waveforms, we report Pearson Correlation ($r$) and Signal-to-Noise Ratio (SNR). For vital signs, we report Mean Absolute Error (MAE) for HR, RR, and the HRV metrics SDNN, RMSSD, and LF/HF.

\paragraph{HRV Calculation Pipeline.}
The estimation of HRV metrics from the predicted PPG waveform follows a multi-stage signal processing pipeline. First, cardiac cycles are identified by detecting valid peaks in the waveform, considering signal prominence, width, and periodicity relative to a rolling frequency estimate. These detections are filtered to retain only high-confidence peaks. From the resulting peak train, Inter-Beat Intervals (IBIs) are calculated. This IBI time-series is then cleaned by interpolating outlier intervals that fall outside a physiologically plausible range (e.g., due to missed or false detections). Finally, standard time-domain (SDNN, RMSSD) and frequency-domain (LF/HF) metrics are computed from the cleaned IBI sequence, contingent on meeting minimum duration (e.g., $\geq$ 20s for SDNN) and beat count thresholds.

\paragraph{Reporting Strategy.}
All results in Section \ref{sec:results} are reported by first averaging metrics for all chunks belonging to a single individual, and then averaging these per-individual scores. This approach prevents individuals with more video chunks from disproportionately influencing the aggregate results. HRV metrics (SDNN, RMSSD) are calculated for all chunks $\geq 20$s. LF/HF, which requires a longer window, is calculated only for chunks $\geq 55$s (approx. 1/3 of the test set).

\section{Results}
\label{sec:results}

\subsection{Overall Performance}

Table \ref{tab:results-main} summarizes the performance of all models on our \num[round-mode=places,round-precision=0]{\nparticipantstest}-participant combined test set. VitalLens 2.0 significantly outperforms all other methods across almost all metrics, establishing a new state-of-the-art for learned rPPG models.

\begin{table}[h!]
  \caption{Vitals estimation results on the combined test set (N=\num[round-mode=places,round-precision=0]{\nparticipantstest} individuals). Results are averaged per-individual. Best performance is in \textbf{bold}.}
  \label{tab:results-main}
  \centering
  \begin{threeparttable}
  \sisetup{round-mode=places}
  {
  \setlength{\tabcolsep}{4pt}
  \begin{tabular}{lccccccccc}
    \toprule
     & \multicolumn{3}{c}{Pulse} & \multicolumn{3}{c}{Respiration\tnote{a}} & \multicolumn{3}{c}{Heart Rate Variability (HRV)} \\
    \cmidrule(r){2-4} \cmidrule(r){5-7} \cmidrule(r){8-10}
     & HR & \multicolumn{2}{c}{PPG} & RR & \multicolumn{2}{c}{RESP} & SDNN & RMSSD & LF/HF\tnote{b} \\
    Method & MAE $\downarrow$ & $r \uparrow$ & SNR $\uparrow$ & MAE $\downarrow$ & $r \uparrow$ & SNR $\uparrow$ & MAE $\downarrow$ & MAE $\downarrow$ & MAE $\downarrow$ \\
    \midrule
    \begin{filecontents*}{data/results_main.csv}
method,hrmae,hrppgcor,hrppgsnr,rrmae,rrrespcor,rrrespsnr,hrvsdnnmae,hrvrmssdmae,hrvlfhfmae
G,12.33349,0.33710,-8.19879,0.00000,0.00000,0.00000,58.11887,81.29211,1.30601
CHROM,3.25682,0.56235,1.86960,0.00000,0.00000,0.00000,30.22656,48.87022,1.11626
POS,4.02508,0.60996,2.68156,0.00000,0.00000,0.00000,50.55899,80.08567,1.35487
DeepPhys,5.17360,0.64610,3.43103,0.00000,0.00000,0.00000,26.57382,40.64781,1.12902
MTTS-CAN,2.17113,0.75554,7.47107,0.00000,0.00000,0.00000,19.58239,31.72709,0.95663
EfficientPhys,2.95963,0.74748,7.15599,0.00000,0.00000,0.00000,19.33648,31.37832,0.98120
VitalLens 1.0*,2.12518,0.80978,9.88295,1.90748,0.74678,7.14960,20.55122,33.17031,1.03402
VitalLens 1.1,1.64312,0.85334,12.88958,1.09114,0.81399,10.09317,12.70066,19.99789,0.85298
VitalLens 2.0,1.56716,0.85688,13.51550,1.08427,0.81616,10.24858,10.18038,16.44912,0.83443
\end{filecontents*}
\csvreader[
        head to column names,
        late after line=\\]
        {data/results_main.csv}
        {
        method=\method,
        hrmae=\hrmae, hrppgcor=\hrppgcor, hrppgsnr=\hrppgsnr,
        rrmae=\rrmae, rrrespcor=\rrrespcor, rrrespsnr=\rrrespsnr,
        hrvsdnnmae=\hrvsdnnmae, hrvrmssdmae=\hrvrmssdmae, hrvlfhfmae=\hrvlfhfmae
        }
        {
         \method &
         \printval{\hrmae}{besthrmae} &
         \printval{\hrppgcor}{besthrppgcor} &
         \printval{\hrppgsnr}{besthrppgsnr} &
         \printval{\rrmae}{bestrrmae} &
         \printval{\rrrespcor}{bestrrrespcor} &
         \printval{\rrrespsnr}{bestrrrespsnr} &
         \printval{\hrvsdnnmae}{besthrvsdnnmae} &
         \printval{\hrvrmssdmae}{besthrvrmssdmae} &
         \printval{\hrvlfhfmae}{besthrvlfhfmae}
        }
    \bottomrule
  \end{tabular}
  }
  \begin{tablenotes}
    \item[a] Calculated on subset containing a ground truth RESP signal (approx. 40\% of test set).
    \item[b] Calculated on subset of chunks $\geq$ 55s in duration (approx. $\frac{1}{3}$ of test set).
    \item[*] VitalLens 1.0 was trained on a smaller dataset.
  \end{tablenotes}
  \end{threeparttable}
\end{table}

Notably, the performance gap is most pronounced in HRV estimation. While retraining the original model architecture on new data (VitalLens 1.1) yields significant improvements over both the original VitalLens 1.0* and other baselines, the new architecture in VitalLens 2.0 provides a further substantial leap in accuracy. For example, VitalLens 2.0 achieves an SDNN MAE of \roundnumdigits{2}{\besthrvsdnnmae} ms.

This demonstrates that the architectural and training methodology improvements in VitalLens 2.0 were critical for achieving the high-fidelity waveform estimation necessary for robust HRV analysis. This result is the basis for enabling HRV metrics only for users of the VitalLens 2.0 model in our API.

\begin{figure}[h!]
  \centering
  \begin{subfigure}{0.32\textwidth}
    \centering
    \begin{tikzpicture}
      \begin{axis}[
          xlabel=Gold-standard HR (bpm),
          ylabel=Estimated HR (bpm),
          height=4.7cm,
          width=\textwidth,
        ]
        \addplot[
          only marks,
          mark=o,
          mark size=.8pt,
          color=red,
          opacity=0.4
        ] table [col sep=comma, x=hr_true, y=hr_est] {data/vl2_hr_scatter.csv};
        \addplot [gray,line width=0.2, domain=40:130, samples=2] {x};
      \end{axis}
    \end{tikzpicture}
    \caption{Heart Rate (HR)}
  \end{subfigure}
  \hfill
  \begin{subfigure}{0.32\textwidth}
    \centering
    \begin{tikzpicture}
      \begin{axis}[
          xlabel=Gold-standard RR (bpm),
          ylabel=Estimated RR (bpm),
          height=4.7cm,
          width=\textwidth,
        ]
        \addplot[
          only marks,
          mark=o,
          mark size=.8pt,
          color=blue,
          opacity=0.4
        ] table [col sep=comma, x=rr_true, y=rr_est] {data/vl2_rr_scatter.csv};
        \addplot [gray,line width=0.2, domain=10:35, samples=2] {x};
      \end{axis}
    \end{tikzpicture}
    \caption{Respiratory Rate (RR)}
  \end{subfigure}
  \hfill
  \begin{subfigure}{0.32\textwidth}
    \centering
    \begin{tikzpicture}
      \begin{axis}[
          xlabel=Gold-standard SDNN (ms),
          ylabel=Estimated SDNN (ms),
          height=4.7cm,
          width=\textwidth,
          xmin=-10, xmax=180,
          ymin=-10, ymax=180,
        ]
        \addplot[
          only marks,
          mark=o,
          mark size=.8pt,
          color=purple,
          opacity=0.4
        ] table [col sep=comma, x=sdnn_true, y=sdnn_est] {data/vl2_sdnn_scatter.csv};
        \addplot [gray,line width=0.2, domain=0:170, samples=2] {x};
      \end{axis}
    \end{tikzpicture}
    \caption{HRV-SDNN}
  \end{subfigure}
  \caption{VitalLens 2.0 estimated vitals vs. gold-standard true vitals on the combined test set.}
  \label{fig:scatterplots}
\end{figure}
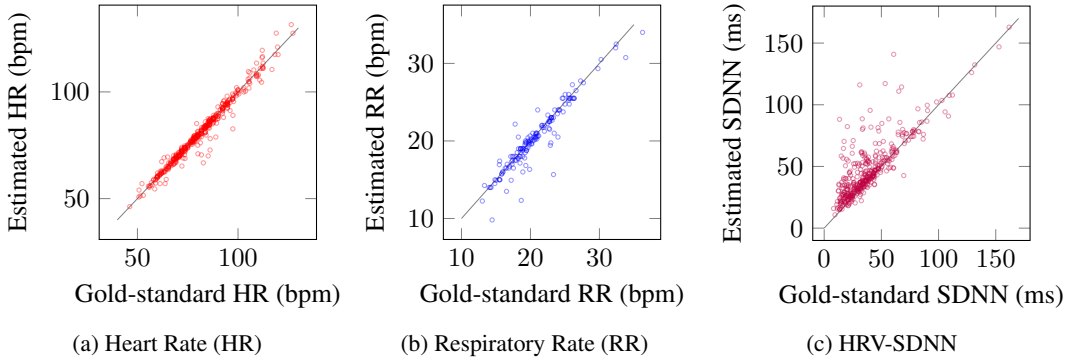

Figure \ref{fig:scatterplots} provides scatter plots for HR, RR, and SDNN, visually demonstrating the high correlation between VitalLens 2.0 estimates and the gold-standard ground truth labels.

\subsection{Results by Dataset}
\label{sec:results-datasets}

To demonstrate the generalization of VitalLens 2.0, Table \ref{tab:results-by-dataset} breaks down performance across all constituent test sets. The results confirm the model's strong performance, and the variance between datasets is explained by their specific, known characteristics. The in-house PROSIT set, which uniquely features unscripted participant and camera motion, serves as our most challenging real-world benchmark. As expected, it shows the highest MAE for HR, RR, and SDNN. Similarly, Vital Videos (Africa) shows a high SDNN MAE of \num{24.77} ms, comparable to PROSIT's. This is also anticipated, as this dataset's primary challenge is its high concentration of participants with Fitzpatrick skin types 5 and 6. In contrast, the Vital Videos (EU) and (South Asia) subsets, which feature stationary subjects and a wider mix of skin tones, demonstrate excellent performance, with SDNN MAE as low as \num{6.00} ms.

\begin{table}[h!]
\caption{VitalLens 2.0 estimation performance by dataset subset. Results are averaged per-individual.}
\label{tab:results-by-dataset}
\centering
\begin{threeparttable}
\sisetup{round-mode=places, round-precision=2}
{
\setlength{\tabcolsep}{6pt}
\begin{tabular}{lccc}
    \toprule
    Source Dataset & HR MAE $\downarrow$ & RR MAE\tnote{a} $\downarrow$ & SDNN MAE $\downarrow$ \\
    \midrule
    
    \begin{filecontents*}{data/results_by_dataset.csv}
source_dataset,hr_mae,rr_mae,sdnn_mae
PROSIT (In-house),3.22300,3.03711,27.45259
Vital Videos (EU),1.29511,1.15467,7.68657
Vital Videos (Africa),1.37343,0.79958,24.76838
Vital Videos (South Asia),0.90215,0.68633,5.99942
UBFC-Phys,2.32834,0.00000,2.81121
UBFC-rPPG,2.53241,0.00000,3.76191
\end{filecontents*}
\csvreader[
        head to column names,
        late after line=\\]
        {data/results_by_dataset.csv}
        {
         source_dataset=\sourceDataset, 
         hr_mae=\hr, 
         rr_mae=\rr, 
         sdnn_mae=\sdnn
        }
        {
         \sourceDataset & \printnum{\hr} & \printnum{\rr} & \printnum{\sdnn}
        }
        
    \bottomrule
\end{tabular}
}
\begin{tablenotes}
    \item[a] RR metrics only given where ground truth labels available.
\end{tablenotes}
\end{threeparttable}
\end{table}

A notable finding comes from the UBFC-Phys and UBFC-rPPG datasets, which show mediocre HR MAE (\num{2.33} and \num{2.53} bpm, respectively) but state-of-the-art SDNN MAE (\num{2.81} and \num{3.76} ms). This apparent discrepancy is likely attributable to the composition of these datasets. The long duration of the UBFC samples (typically 60 seconds) provides a highly stable window for time-domain HRV analysis, which is less sensitive to the challenges that affect frequency-domain HR estimation. Furthermore, the higher average heart rates in these datasets (approx. 90 bpm) may present a greater challenge for frequency-based HR algorithms. Finally, the RR MAE results reinforce these themes, with the high-motion PROSIT set showing significantly higher error (\num{3.04} bpm) than the stationary Vital Videos subsets. This confirms that the core challenges identified, namely participant motion and dataset composition, are consistent across all estimated vital signs \cite{rouast2023vitallens}.

\subsection{Robustness Analysis}
\label{sec:results-factors}

\begin{figure}[h!]
    \centering
    \begin{tikzpicture}
        \begin{groupplot}[
            group style={
                group size=2 by 1,
                horizontal sep=2.5em,
                vertical sep=1em
            },
            height=4.6cm,
            width=0.49\textwidth,
            table/col sep=comma,
            ybar,
            grid=major,
            xtick=data,
            ymin=0
        ]
        \nextgroupplot[
            ylabel=SDNN MAE (ms),
            xlabel={Participant Movement},
            xticklabels from table={data/sdnn_movement.csv}{bin},
            bar width=18pt, enlarge x limits=0.3,
            legend to name=grouplegend,
            legend style={legend columns=-1, draw=none, fill=none, font=\small}
        ]
        \addplot+[blue!60, fill=blue!60] table[x expr=\coordindex, y=vl2_mae] {data/sdnn_movement.csv};
        \addlegendentry{VitalLens 2.0}
        \addplot+[red!60, fill=red!60] table[x expr=\coordindex, y=vl1_mae] {data/sdnn_movement.csv};
        \addlegendentry{VitalLens 1.0*}
        \node[anchor=north west, font=\bfseries] at (rel axis cs:0.05,0.95) {(a)};
        \nextgroupplot[
            xlabel={Participant Skin Type},
            xticklabels from table={data/sdnn_skin_type.csv}{bin},
            bar width=8pt, enlarge x limits=0.15
        ]
        \addplot+[blue!60, fill=blue!60] table[x expr=\coordindex, y=vl2_mae] {data/sdnn_skin_type.csv};
        \addplot+[red!60, fill=red!60] table[x expr=\coordindex, y=vl1_mae] {data/sdnn_skin_type.csv};
        \node[anchor=north west, font=\bfseries] at (rel axis cs:0.05,0.95) {(b)};
        \end{groupplot}
        \node (legend) at ($(group c1r1.north)!0.5!(group c2r1.north)$) [yshift=1.2em] {\ref*{grouplegend}};
    \end{tikzpicture}
    \caption{Comparing the robustness in HRV-SDNN estimation between VitalLens 2.0 and VitalLens 1.0* under increasing participant movement and different participant skin types.}
    \label{fig:sdnn-robustness}
\end{figure}
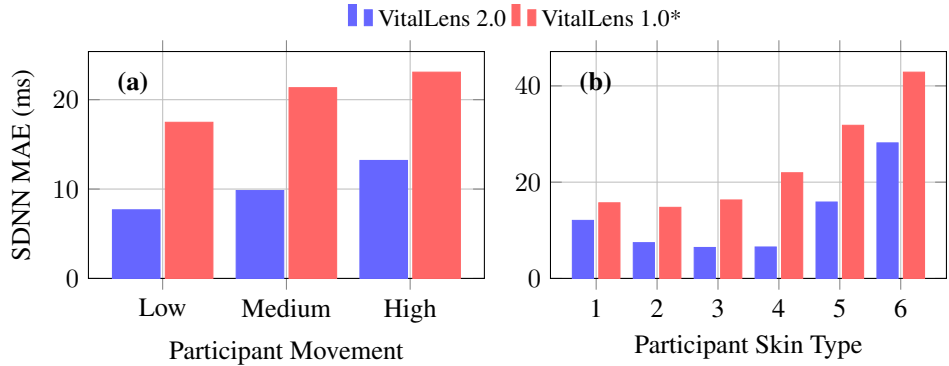

The dataset-level analysis in Section \ref{sec:results-datasets} demonstrates that performance is dictated by specific, known challenges. The high error rates in the PROSIT and Vital Videos (Africa) subsets, for example, directly point to participant movement and darker skin tones as the two most critical factors for robust, real-world HRV estimation. To isolate and quantify the impact of these factors, we conducted this more granular analysis, comparing VitalLens 2.0 against the previous VitalLens 1.0* model. The results in Figure \ref{fig:sdnn-robustness} not only confirm this hypothesis but also highlight the practical value of the new architecture.

Figure \ref{fig:sdnn-robustness}(a) illustrates the model's performance against participant movement, a critical factor for any non-clinical or handheld application. We binned the test set into terciles (Low, Medium, and High) based on a computed motion metric. The results clearly show that while motion artifacts remain a challenge for both models, with error rates increasing in line with motion, the new architecture in VitalLens 2.0 provides a substantial, systematic improvement. The key finding is a clear downward shift in absolute error across all three bins. For instance, we observe that the SDNN MAE for VitalLens 2.0 in the 'High' movement category is lower than the error for VitalLens 1.0* in the 'Low' movement category.

Similarly, Figure \ref{fig:sdnn-robustness}(b) analyzes performance across all six Fitzpatrick skin types. The data reveals two key findings. First, VitalLens 2.0 delivers a new level of performance, achieving a low and highly stable error rate across skin types 1 through 4, in contrast to the rising error of VL 1.0*. Second, for skin types 4, 5, and 6, where the rPPG signal is most attenuated, VitalLens 2.0 provides a strong reduction in error. This confirms the effectiveness of our expanded and diversified training dataset, which included a significant number of individuals with darker skin tones. However, the results also transparently show that a performance gap remains; error rates for types 5 and 6 are still notably higher than for types 1-4. While VL 2.0 makes HRV estimation more equitable and reliable, further research is required. Much as our previous work focused on stabilizing HR estimation across all skin tones, achieving this same equity for high-fidelity HRV metrics represents the new frontier, and it remains a key focus for our ongoing development.

\section{Conclusion}
\label{sec:conclusion}

VitalLens 2.0 represents a significant advancement in remote physiological monitoring. By combining a large-scale, diverse dataset with a novel, optimized architecture, it achieves state-of-the-art accuracy not only for heart and respiration rates but also for challenging Heart Rate Variability metrics. The demonstrated high-fidelity performance, particularly for HRV, opens new possibilities for accessible, non-invasive health and wellness tracking.

\begin{ack}
We are thankful to Pieter-Jan Toye for his helpful suggestions and access to the Vital Videos dataset. This work was funded by Rouast Labs Pty Ltd.
\end{ack}

\bibliographystyle{plain}
\bibliography{paper} 


\end{document}